%% file: main.tex
\documentclass[10pt,twocolumn,letterpaper]{article}

 \usepackage{wacv}              

\definecolor{wacvblue}{rgb}{0.21,0.49,0.74}
\usepackage[pagebackref,breaklinks,colorlinks,allcolors=wacvblue]{hyperref}
\input{preamble}
\def\wacvPaperID{423} 
\def\confName{WACV}
\def\confYear{2027}
\usepackage[accsupp]{axessibility}  

\title{Frequency-Adaptive Gaussian Grids for Efficient Image
Representation}

\author{
Anatolii Evdokimov$^{1}$ \quad
Stavros Diolatzis$^{2}$ \quad
Bingxuan Li$^{1}$ \quad
Akshay Jindal$^{2}$ \quad \\
Patsorn Sangkloy$^{1}$ \quad
Qi Sun$^{1}$
\\[0.5em]
$^{1}$New York University \qquad $^{2}$Intel Corporation
\\[0.3em]
{\tt\small \{tolya.e, bingxuan.li, ps5688, qisun\}@nyu.edu}
{\tt\small \{stavros.diolatzis, akshay.jindal\}@intel.com}
}

\begin{document}

\maketitle
\input{conf/symbols}
\input{content/figures/teaser}
\input{content/frontmatter/abstract}
\input{content/main/1-introduction}
\input{content/main/2-related}
\input{content/main/3-method}
\input{content/main/4-result}
\input{content/main/5-future-work}

{
    \small
    \bibliographystyle{ieeenat_fullname}
    \bibliography{bibliography}
}

\end{document}

%% file: conf/symbols.tex
\newcommand{\projName}{Gaussian-Grid\xspace}

\newcommand{\fftWindowSize}{w}

\newcommand{\pixI}{i}
\newcommand{\pixJ}{j}
\newcommand{\Image}{I}
\newcommand{\FCidx}{\mathbf{u}}
\newcommand{\FFT}{\mathcal{F}}

\newcommand{\numGaus}{N}
\newcommand{\GausFun}{\mathcal{G}}
\newcommand{\gausSub}{n}
\newcommand{\gausMu}{\mu}
\newcommand{\gausSigma}{\sigma}
\newcommand{\gausTheta}{\theta}
\newcommand{\gausRotMat}{R}
\newcommand{\gausCovMat}{\Sigma}
\newcommand{\gausPixCoord}{\mathbf{x}}
\newcommand{\gausFreq}{f}
\newcommand{\freqPixel}{r}
\newcommand{\topk}{K}
\newcommand{\tile}{T}
\newcommand{\overlapGaus}{\mathcal{S}}
\newcommand{\tileSub}{l}

\newcommand{\Grid}{grid}
\newcommand{\Decoder}{MLP\textsubscript{dec }}
\newcommand{\CanvasGrid}{canvas grid }

\newcommand{\Embed}{\textbf{v}}
\newcommand{\PE}{\gamma}
\newcommand{\cornerFeatures}{\textbf{f}_{corner}}
\newcommand{\gridCellX}{x}
\newcommand{\gridCellY}{y}
\newcommand{\levelsCanv}{\textit{L}\textsubscript{canv}}
\newcommand{\levelsComp}{\textit{L}\textsubscript{comp}}
\newcommand{\finalSignal}{\textbf{c}\textsubscript{r}}

\newcommand{\ntcName}{\textbf{GRID}\xspace}

\newcommand{\bitDepth}{\textit{B}}
\newcommand{\scalingFac}{s}
\newcommand{\tensor}{\textit{X }}
\newcommand{\uniform}{\textit{U}}
\newcommand{\noise}{n}
\newcommand{\quantized}{q}

\newcommand{\first}{\cellcolor{blue!25}\bfseries}
\newcommand{\second}{\cellcolor{blue!10}}

%% file: content/figures/teaser.tex
\begin{figure*}[t]
   \includegraphics[width=0.34\textwidth]{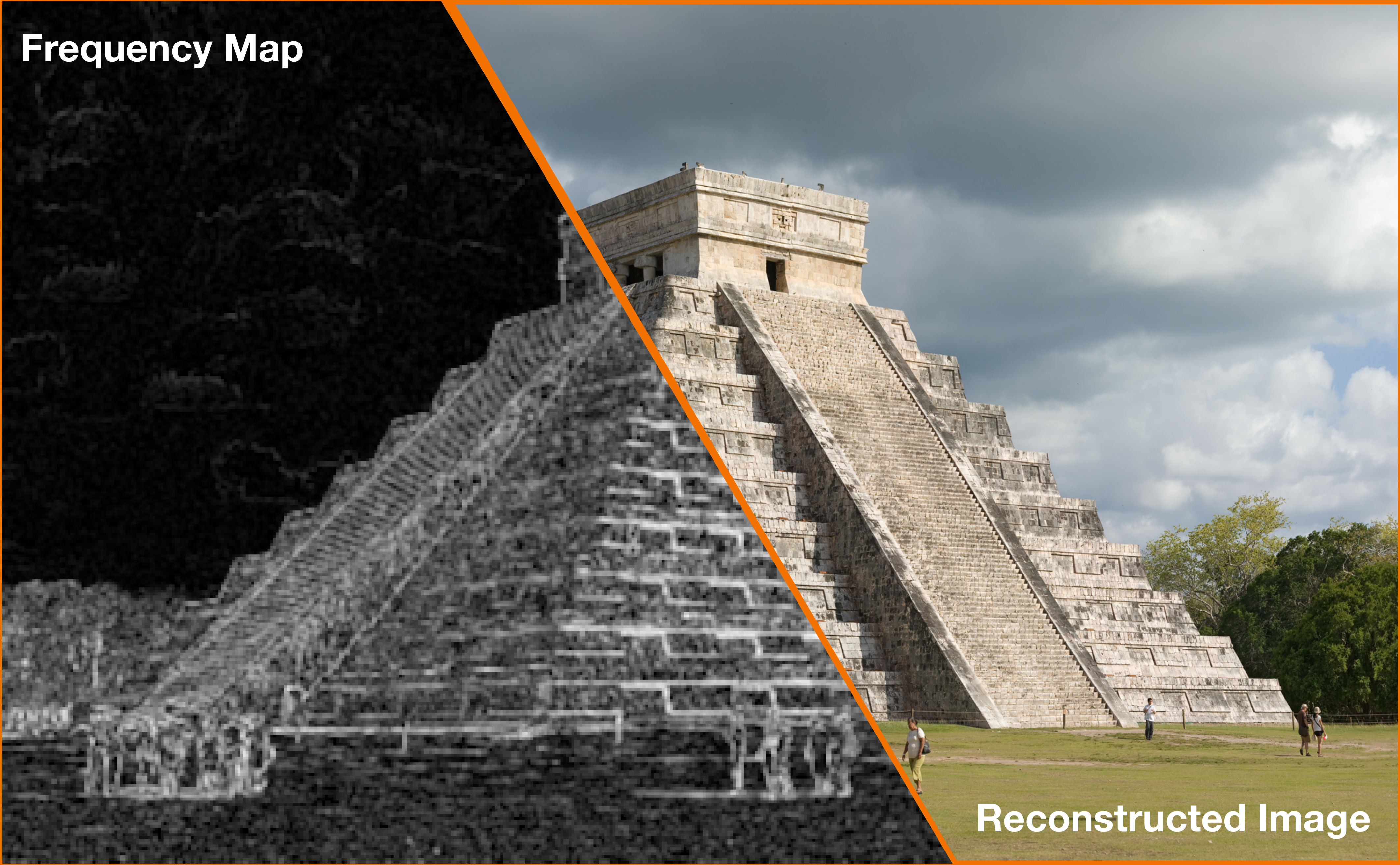}
   \includegraphics[width=0.34\textwidth]{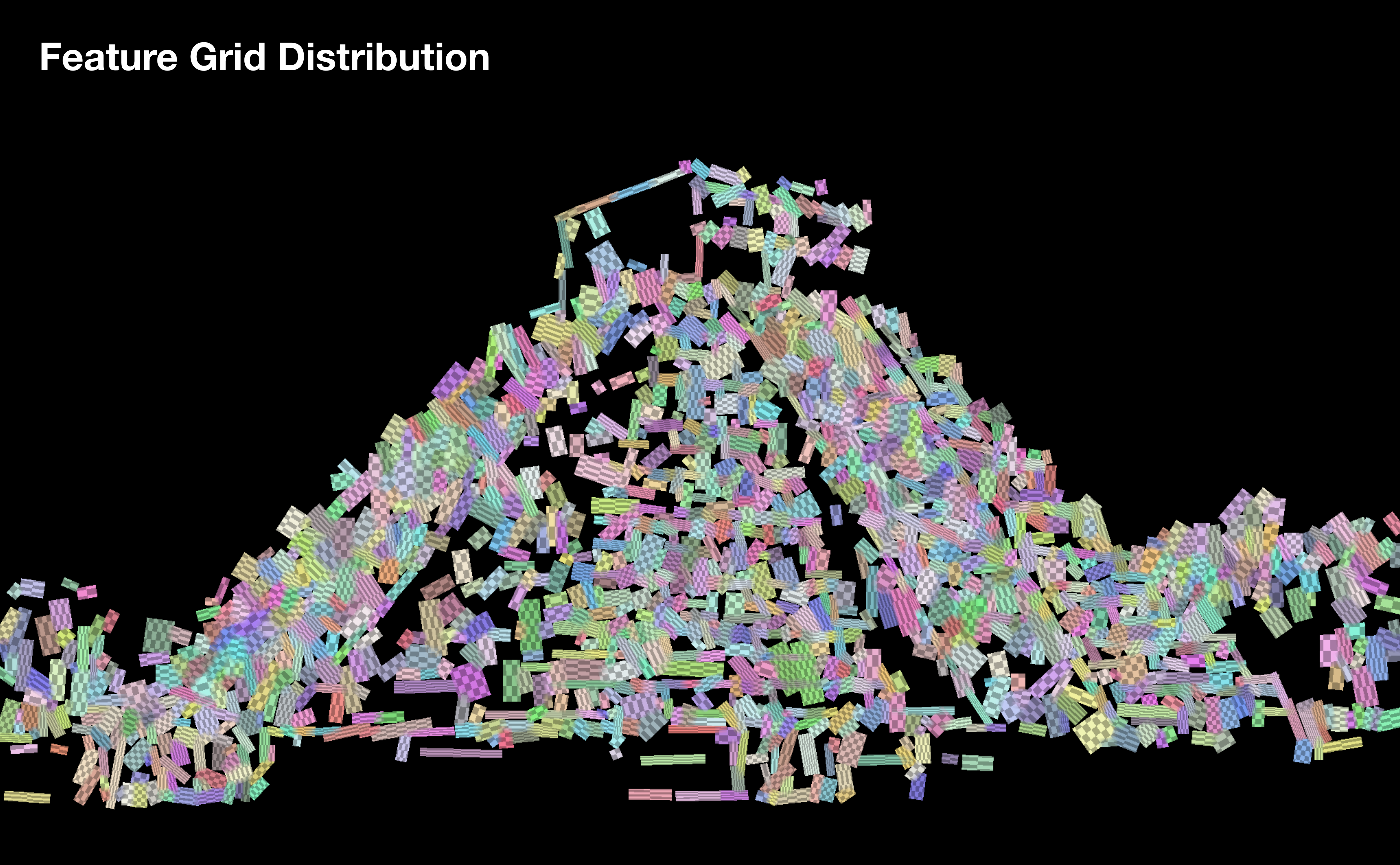}
   \includegraphics[width=0.29\textwidth]{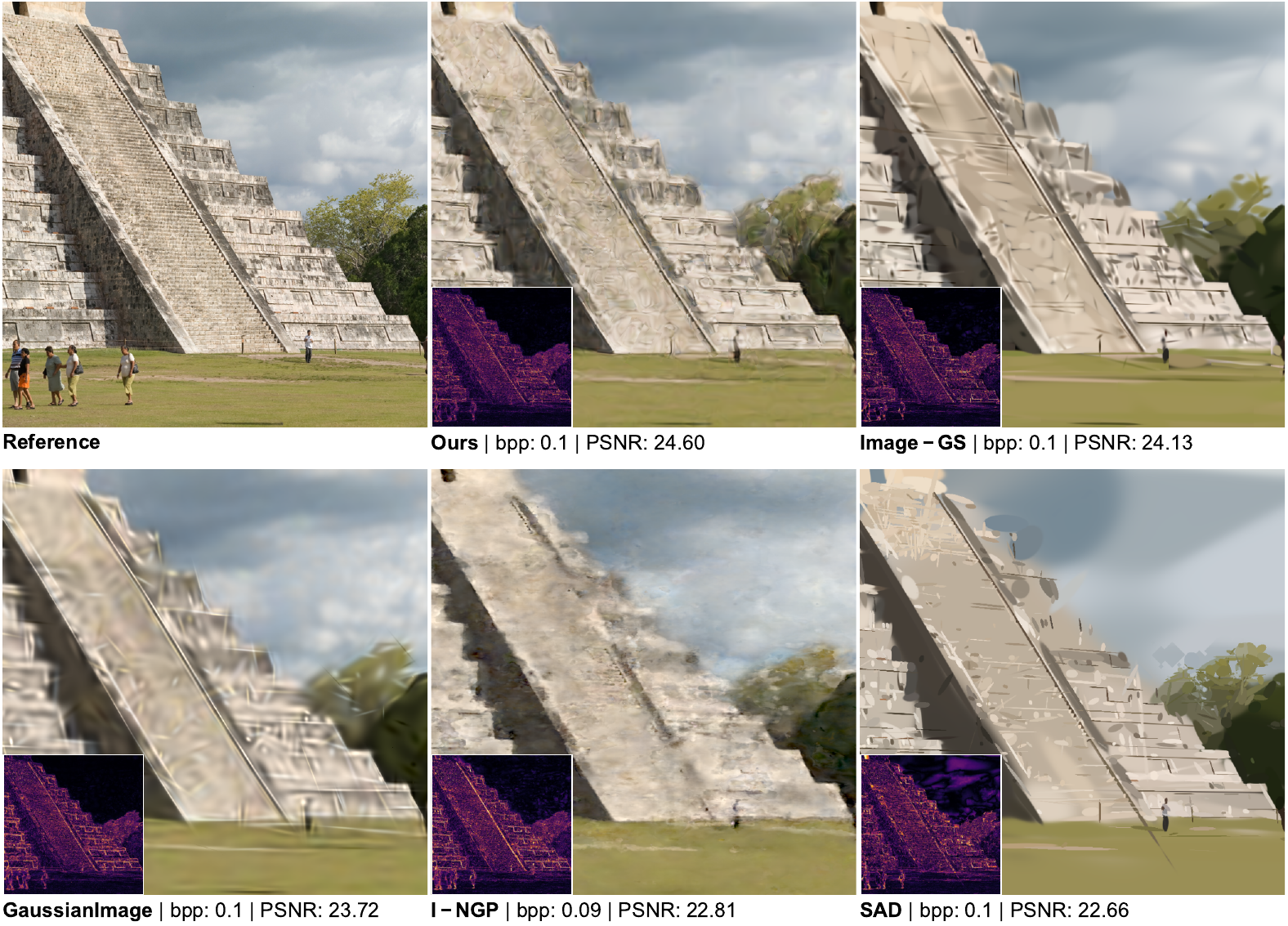}
  \caption{Gaussian-Grid leverages Gaussian-defined feature grid layout that is guided by image frequency map to generate high quality reconstruction.} 
  \label{fig:teaser}
\end{figure*}

%% file: content/frontmatter/abstract.tex
\begin{abstract}
Neural representation has emerged as a powerful tool to mitigate the demand for increasingly higher-resolution images. However, a fundamental trade-off remains between structured (e.g., feature grids, which offer spatial consistency) and unstructured (e.g., 2D Gaussians, which adaptively capture local variances) representations. This choice significantly affects how well an algorithm represents an image. Here, we present \projName, a hybrid neural representation that adaptively integrates structured and unstructured primitives. By dynamically partitioning computation between 2D Gaussian distributions and their attached multi-resolution feature grids based on the local frequency content, \projName automatically optimizes its architecture to match the underlying complexity of the target signal. We evaluate our approach using high-resolution natural images and demonstrate its applications as GPU-native dataset storage, showing $1.24$--$2.14\times$ decoding speed improvement and PCIe data traffic reduction of $23\times$--$176\times$ over JPEG + ZIP data archiving. The results show that \projName maintains small memory footprints while achieving high reconstruction fidelity across a variety of datasets.
\end{abstract}

%% file: content/main/1-introduction.tex
\section{Introduction}
\label{sec:introduction}

Representing images with high fidelity and low memory footprint is foundational in GPU systems and computer vision applications, especially with the recent surging demand for ultra-high-quality image data. Traditional grid-based representations scale quadratically in memory, creating a bottleneck where discrete sampling rates fail to maintain visual quality due to storage and bandwidth constraints.

To tackle the scaling limitations, neural representations have emerged over the past decade as an efficient alternative by learning image signals using neural networks. 
Among these approaches, images and texture stacks may be represented through a combination of structured feature grids \cite{vaidyanathan2023random,martel2021acorn,muller2022instant} and neural networks to represent localized spatial signals. 
Alternatively, unstructured explicit primitives such as 2D Gaussians \cite{zhang2025image,zhang2024gaussianimage} enable spatially adaptive representations that optimize resource allocation to where it is warranted. 
However, both directions have pros and cons: structured grids prioritize memory efficiency by implicitly defining feature distributions but suffer from limited spatial flexibility, whereas primitive-based methods offer better adaptivity at the cost of higher overhead for complex local details.

Here, we introduce \projName, a hybrid representation that integrates both structured and unstructured neural representations. By unifying the spatial flexibility of primitives with the efficient representation power of feature grids, \projName achieves a balance between fidelity and memory efficiency for high-resolution imagery across a wide range of content types.

Our main insight is to represent images as collections of 2D Gaussians that each carry an embedded feature grid. In this architecture, the Gaussians act as a substrate, optimizing for global geometric structures and automatically anchoring high-capacity feature grids only where local signal complexity demands them.

To this aim, we design a two-stage algorithm composed of a frequency-guided framework to decide the Gaussian distribution and a final grid optimization step. The transition from the frequency-guided Gaussians to the grid optimization happens by attaching feature grids of appropriate resolution based on the local complexity represented by each Gaussian. Finally, these features are decoded through a lightweight multilayer perceptron (MLP).

We evaluate \projName across a variety of realistic and high-resolution images adapted from DIV2K \cite{Agustsson2017DIV2K} and CLIC2020 \cite{CLIC2020}. The results demonstrate that \projName robustly handles images of varying frequency content due to its ability to switch from a structured to an unstructured representation. As such, our method outperforms structured neural representations for concentrated frequency content and primitive based representations for uniform frequency content. We further demonstrate that our method serves as a GPU-native image dataset archiver that bypasses the CPU-dependent decompression overhead of popular ZIP and JPEG solutions, achieving $1.24$--$2.14\times$ decoding speed improvement and PCIe data traffic reduction of $23\times$--$176\times$ over JPEG + ZIP data archiving. Our source code is released at \url{https://github.com/NYU-ICL/gaussian-grid}.

%% file: content/main/2-related.tex
\section{Related Work}
\label{sec:related}
Prior work on image representation can be broadly categorized into traditional block-based compression, structured neural representations based on regular grids, and unstructured neural representations that use continuous functions or learned primitives. These approaches exhibit different trade-offs between spatial adaptivity, memory efficiency, and decoding cost. Our work draws inspiration from all three lines while addressing their respective limitations in representing spatially heterogeneous high-frequency content.

\subsection{Traditional Representations}
Traditional techniques are designed to satisfy the constraints of real-time graphics hardware, including fast decoding, low memory bandwidth, and constant-time random access \cite{akenine2019real}. As a result, most formats adopt fixed-rate, block-based representations that compress small spatial regions independently. Early methods such as block truncation coding \cite{textureBTC1979} introduced local approximation using a small number of representative values per block, an idea later extended to color textures \cite{campbell1986two} and standardized in widely deployed GPU compression formats. Subsequent formats added support for alpha channels, normal maps, and high dynamic range content \cite{munkberg2006high,roimela2006high}, while more recent standards such as ASTC allow variable block sizes and bitrates \cite{strom2005packman,strom2007etc,fenney2003texture,nystad2012adaptive}. These designs improve compression quality and flexibility but remain constrained by fixed block layouts and minimum per-block bitrates, which cap achievable compression ratios. Moreover, capacity is distributed uniformly across blocks, preventing adaptation to spatially varying signal complexity. High-frequency details and sharp transitions are forced to share the same bit budget as smooth regions, leading to inefficient allocation and visible artifacts at low bitrates. These limitations motivate neural representations that move beyond block-based compression and enable more flexible trade-offs between memory usage, reconstruction quality, and spatial capacity allocation.

\subsection{Structured Neural Representation}
Structured neural representations encode signals using spatially aligned feature grids coupled with lightweight neural decoders, enabling efficient storage and fast evaluation. Early work demonstrated that dense and multi-resolution grids can significantly reduce memory usage while maintaining high reconstruction quality \cite{chen2021learning,martel2021acorn}. Subsequent methods introduced hash-based and learned indexing schemes and support for discontinuities to further improve scalability and inference performance \cite{muller2022instant, takikawa2022variable, takikawa2023compact, belhe2023discontinuity}.

These approaches have recently been applied to texture compression and appearance representation \cite{takikawa2023compact}. Neural Texture Compression (NTC) \cite{vaidyanathan2023random}, for example, encodes multi-channel textures and mipmap chains using uniform feature grids and a lightweight decoder, achieving strong compression ratios and fast random access. Extensions include block-compressed features \cite{weinreich2024real} and differentiable indirection structures \cite{datta2023efficient} to improve scalability for large textures.

Despite their efficiency, structured grid-based representations allocate capacity uniformly across space. Localized high-frequency details are therefore supported everywhere in the grid, resulting in over-allocation in smooth regions and limited adaptation to spatially heterogeneous frequency distributions. As a consequence, pure grid-based methods are less effective for high-resolution textures with concentrated or non-uniform detail.

\subsection{Unstructured Neural Representation}
Unstructured neural representations depart from spatially regular parameterizations and instead model signals as continuous functions. Coordinate-based implicit neural representations (INRs) map spatial coordinates directly to signal values using multilayer perceptrons with positional or sinusoidal encodings \cite{sitzmann2019siren,dupont2021coincompressionimplicitneural, dupont2022coinneuralcompressionmodalities,wang2023implicit,shenouda2024relus}. These methods are highly expressive and resolution-independent but often suffer from slow inference due to dense per-query network evaluation and limited scalability for large images or multi-channel textures.

An alternative class of unstructured representations uses explicit learnable primitives, most notably Gaussian functions. Gaussian-based methods represent images as collections of localized primitives parameterized by position, shape, and appearance, and reconstruct signals by aggregating their contributions \cite{zhang2024gaussianimage, zeng2025instant,zhang2025image}. An alternative method using soft anisotropic diagrams as a primitive has also been recently proposed \cite{iinbor2026sad}. By adjusting primitive density and size, these approaches can adaptively allocate capacity to complex regions while keeping smooth areas compact.
While Gaussian primitives offer strong spatial adaptivity, each primitive introduces non-trivial parameter overhead. For dense textures or high-capacity representations, the total number of primitives required to model fine details can grow rapidly, reducing efficiency. This trade-off motivates hybrid approaches that combine the adaptive placement of unstructured primitives with the compactness and regularity of structured feature grids.

\subsection{Hybrid Neural Representations}
Hybrid neural representations combine structured parameterizations with unstructured or continuous components to balance efficiency and expressiveness. ACORN \cite{martel2021acorn} introduces content‑adaptive grid density through a hierarchical block decomposition coupled with a lightweight decoder, demonstrating that allocating finer grids to complex regions improves large‑scale image representation. However, ACORN relies on axis‑aligned grids, limiting its efficiency for anisotropic or oriented structures such as edges and textured patterns.

In 3D reconstruction, recent hybrid approaches augment explicit primitives with spatially varying detail. Textured Gaussian methods \cite{chao2025textured, wang2025neural} extend 3D Gaussian Splatting by attaching local texture maps to individual Gaussians, enabling high‑frequency appearance modeling within each primitive’s support. Related frequency‑adaptive NeRF \cite{zhang2025lookcloser,rong2025nexels} and Gaussian‑based methods \cite{papantonakis2025content} similarly allocate capacity based on local scene complexity. While these ideas align with our goal of distributing capacity according to signal frequency, they have largely been explored in 3D settings. Their application to purely 2D image representation and compression remains underexplored. To address this gap, \projName combines frequency‑adaptive Gaussian placement with localized feature grids, enabling fine‑grained capacity allocation while maintaining efficient storage and decoding for high‑resolution images.

%% file: content/main/3-method.tex
\section{Method}
\label{sec:method}
\input{content/figures/method_overview}
The core contribution of our method is a hybrid representation that integrates structured feature grids with unstructured Gaussian primitives. We achieve this by adaptively adjusting representation capacity based on the local frequency content of an image, allocating more parameters to high-frequency areas and fewer to low-frequency ones.

Specifically, we optimize learnable 2D Gaussians to match a local frequency map derived from the target image. These Gaussians provide the spatial layout for attaching high-capacity feature grids only where they are needed. For areas with lower frequency, a \CanvasGrid reconstructs the smooth, low-frequency content. Finally, an MLP decoder (\Decoder) decodes the latent grid features per image. \Cref{fig:method_overview} shows an overview of our proposed method.

\subsection{Gaussian Frequency Distribution Map}
\input{content/figures/rendering_overview}
First, we establish a ground truth frequency distribution to guide the optimization of the Gaussians.
For each spatial window of size $8 \times 8$ centered at pixel $(\pixI, \pixJ)$, we compute the local frequency value $F(\pixI, \pixJ)$ as 

\begin{equation}
    F(\pixI, \pixJ) = \max_{\FCidx \neq 0} \left| \FFT \left[\Image_{(\pixI, \pixJ)} \right] (\FCidx) \right|, 
\end{equation}

\noindent where $\Image_{(\pixI, \pixJ)}$ is the windowed image patch , $\FFT$ denotes the 2D discrete Fourier transform (DFT), $\FCidx$ indexes the frequency components. The DC term is excluded from the maximum calculation to ensure invariance to local brightness variations.

For Step 1 of our optimization, we represent this frequency map as a set of \numGaus{} learnable 2D anisotropic Gaussians. We parametrize each Gaussian $\GausFun_{\gausSub}$ by its center $\gausMu_{\gausSub}$, a diagonal covariance matrix $\gausCovMat_{\gausSub} = diag(\gausSigma_{(\pixI, \gausSub)}, \gausSigma_{(\pixJ, \gausSub)})$, a scalar frequency assignment $\gausFreq_{\gausSub}$, and a rotation angle $\gausTheta_{\gausSub}$. The Gaussian distribution of the $\gausSub$-th Gaussian at coordinate $\gausPixCoord = [\pixI, \pixJ]$ is defined by the equation:
\begin{equation}
    \GausFun_{\gausSub}\left(\gausPixCoord\right) = \exp\left(-\frac{1}{2}(\gausPixCoord - \gausMu_{\gausSub})^{T} \gausRotMat_{\gausSub} \gausCovMat_{\gausSub}^{-1} \gausRotMat_{\gausSub}^{T}(\gausPixCoord - \gausMu_{\gausSub})\right),
\end{equation}

\noindent where $\gausRotMat_{\gausSub}$ is the 2D rotation matrix for angle $\gausTheta_{\gausSub}$.

The final pixel value $\gausFreq_{\freqPixel}$ is rendered by accumulating each Gaussian's frequency assignment contributions for each coordinate $\gausPixCoord$. Specifically, for each non-overlapping tile $\tile_{\tileSub}$, we determine which Gaussians overlap with the tile in a 3-standard-deviation enclosing circle range (denoted as $\overlapGaus_{\tileSub}$) and render only the contributions of the overlapping Gaussians for that $\tile_{\tileSub}$. This allows us to efficiently consider all Gaussians overlapping a tile.

The final formula for Gaussian frequency map generation is:
\begin{equation}
   \gausFreq_{\freqPixel} = \sum_{n \in \overlapGaus_{\tileSub}(\gausPixCoord)} \GausFun_{\gausSub}(\gausPixCoord) \cdot \gausFreq_{\gausSub}.
\end{equation}
\Cref{fig:method_overview} includes an example visualization of the frequency and Gaussian maps. 

\subsection{Grid Resolution and Placement}
Once the Gaussians are optimized to match the frequency distribution of the given image, we switch to the grid optimization (Step 2). We place grids of appropriate resolution to match the position and shape of each Gaussian. Specifically, the grid budget is determined using the learned $f_n$ value and the pixel coverage of each Gaussian. We compute the weight $w_{\gausSub}$ for each Gaussian grid that determines the proportion of the overall bit budget assigned to that grid:
\begin{equation}
    w_{\gausSub} = \frac{|f_n|}{\text{max}_{j}|f_j|} \cdot 4\gausSigma_{x,n} \cdot 4 \gausSigma_{y,n}
\end{equation}

\noindent where $\gausSigma_{x,n}$ and  $\gausSigma_{y,n}$ are the Gaussian scales (the factor of 4 gives a bounding box of $\pm2\gausSigma$). This way, larger Gaussians with higher frequency magnitude $f_n$ receive more features, while lower-complexity regions receive fewer. We then refine this allocation using directional analysis of the local frequency field.

For each Gaussian, we gather the per-pixel frequency vectors $\mathbf{v}_p = (F_x(\pixI, \pixJ), F_y(\pixI, \pixJ))$ assigned to it and compute their magnitude-weighted directional variance:
\begin{equation}
\mathrm{Var}_{\gausSub} = \frac{\sum_p \|\mathbf{v}_p\| \left\| \hat{\mathbf{v}}_p - \bar{\mathbf{d}}_{\gausSub} \right\|^2}{\sum_p \|\mathbf{v}_p\|},
\end{equation}
\noindent where $\hat{\mathbf{v}}_p = \mathbf{v_p} / ||\mathbf{v_p}||$ is the direction and $\bar{\mathbf{d}}_{\gausSub}$ is the normalized mean direction for Gaussian $\GausFun_{\gausSub}$. We increase the budget for Gaussians with mixed directional content and reduce it for regions with simple, consistent directionality. For each Gaussian, the assigned grid cell count is finally split into height and width according to Gaussian anisotropy, with an additional directional aspect-ratio correction.

In addition to local Gaussian grids, we include a global \CanvasGrid spanning the full image to provide stable low-frequency background coverage. Its budget proportion is 30\% by default but adapted per image via a score $s \in [0, 1]$ derived from the coefficient of variation of $F$ and a global edge term (see supplemental for the full formula). Images with flat, homogeneous content score $s \approx 1$ and receive a larger canvas budget and heavier Gaussian pruning; textured images score $s \approx 0$ and retain dense Gaussian coverage with a smaller canvas budget.

During rendering, each pixel queries its top-$K$ Gaussian grids, sampling features from each local grid in the Gaussian's own coordinate frame. Samples outside a grid are marked invalid, and the final feature is
blended from valid contributors only.

\subsection{Grid Sampling and Interpolation}

For every Gaussian selected by top-K, we bilinearly sample the corresponding local feature grid. The per-Gaussian sampled features are then fused by normalized masked averaging across the valid top-K contributors. The global \CanvasGrid is also sampled for every pixel and added as a low-frequency background component during the blending stage. The blended feature vector is finally passed through a small 2-layer \Decoder to predict the output signal. ~\Cref{fig:rendering_overview} summarizes our rendering pipeline.

\subsection{Quantization}
To achieve high representation compactness while maintaining signal fidelity, our model employs a quantization-aware training strategy (QAT) inspired by NTC \cite{vaidyanathan2023random}. During training, both grid features and \Decoder computations are passed through a simulated quantization pipeline. 

For a given feature grid tensor and a target bit-depth \bitDepth, we first compute a per-tensor scaling factor $\scalingFac = \frac{\text{max}(|\tensor|)}{2^{\bitDepth}-1}$. The floating-point tensor \tensor is then scaled yielding the quantized tensor $\tensor_{\quantized}$. We also add uniform random noise $\noise$ in the range $\noise \sim \uniform(-0.5, 0.5)$:
\begin{equation}
    \tensor_{\quantized} = \text{round}(\frac{\tensor}{\scalingFac} + \noise)
\end{equation}
\noindent The resulting values are clamped to the integer range $[-2^{\bitDepth - 1}, 2^{\bitDepth - 1} - 1]$. 

For the \Decoder, we apply a per-channel quantization scheme to its weights, where a separate scale is calculated for each output channel of the linear layers. The QAT procedure is applied during the first 95\% of the training time, while for the last 5\%, the grid features are hard-quantized and frozen, allowing the \Decoder to adapt to fully quantized low-bit grids.

\subsection{Implementation Details}
\projName is trained in a two-stage process: first, the 2D Gaussian optimization and, second, the grid optimization.

In the first stage, we set learning rates for ($\gausMu, \gausSigma, \gausTheta, \gausFreq$) to (1e-2, 1e-3, 1e-3, 1e-2). The number of initial Gaussians is set per budget: 800, 1500, 2000,
and 2300 for 0.1, 0.4, 0.7, and 1.0\,bpp on DIV2K, and 400, 900,
1200, and 1500 on CLIC2020. The loss function for the first stage is:

\begin{equation}
\mathcal{L}_1 = \mathcal{L}_{L1} + \lambda_{\text{ov}} \cdot \mathcal{L}_{\text{ov}},
\end{equation}

\noindent where $\mathcal{L}_{L1}$ is the L1 loss between the predicted and the target frequency maps and the overlap penalty $\mathcal{L}_{\text{ov}}$ discourages large spatial overlap between Gaussians. For each pair of Gaussians $i$ and $j$, we compute a Gaussian overlap using the pairwise Mahalanobis distance under the summed covariances:

\begin{equation}
\mathcal{L}_{\text{ov}} = \sum_{i < j} \frac{1}{2\pi} \exp\!\left(
    -\frac{1}{2}
    (\gausMu_i - \gausMu_j)^\top
    (\gausCovMat_i + \gausCovMat_j)^{-1}
    (\gausMu_i - \gausMu_j)
\right),
\end{equation}

\noindent where $\gausCovMat$ is the full covariance matrix of a Gaussian. This penalty is summed over all unique pairs and encourages Gaussians to spread apart and tile the image without excessive redundancy. We set the loss weight $\lambda_{\text{ov}} = 5 \times 10^{-5}$.

After the optimization of 2D Gaussians, we begin training the feature grids against the RGB image using an L2 loss between the predicted color and the ground truth. Each pixel queries its top-$K{=}5$ grids, sampling features from each local grid in the Gaussian's coordinate frame. The decoder \Decoder has 2 hidden layers of width 64 and each grid feature has dimension 6, with learning rates of 1e-3 and 1e-2 respectively. We apply 8-bit quantization to \Decoder and 4-bit quantization to the grids, and hard-quantize the Gaussian parameters to 16-bit floats at the start of the second stage. For both stages we use the Adam optimizer~\cite{kingma2017adam}.

To ensure efficient training we implement performance-critical components using GPU kernels. The first-stage Gaussian optimization uses Taichi~\cite{taichi2019}, while the second-stage reconstruction uses a custom CUDA kernel. Gaussians are optimized for 3K steps and the feature grids for 75K steps, yielding an average training time of 24.3 minutes per image on a single A100 GPU.

%% file: content/figures/method_overview.tex
\begin{figure*}[t]
  \centering
  \includegraphics[width=\linewidth,trim={0 0.2cm 0 0cm}]{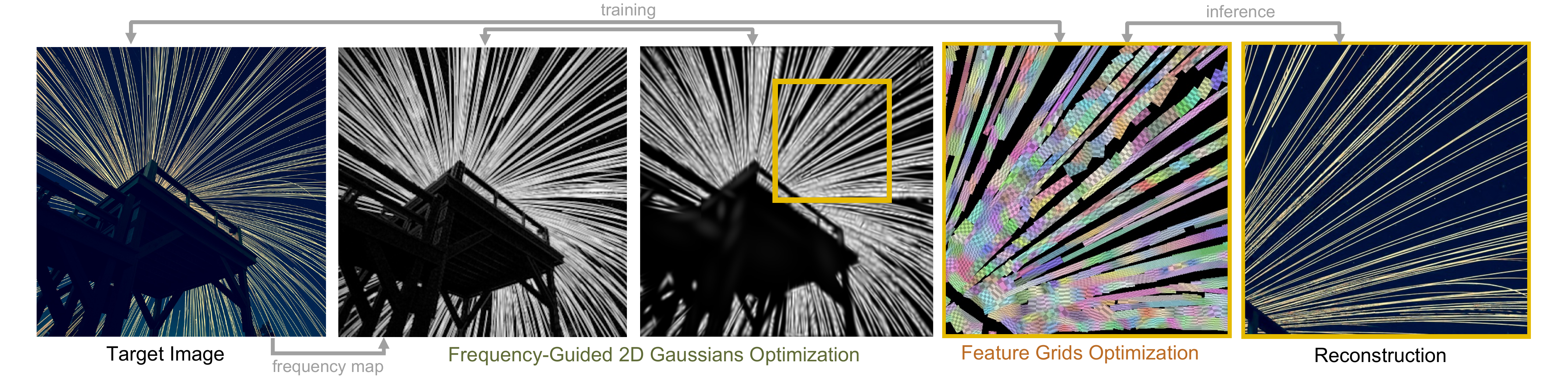}
  \caption{
      Overview of our method. \projName optimizes 2D Gaussian primitives to a spatial frequency map derived from the target image. Gaussians are then used to place feature grids of variable resolutions, aligning with the underlying frequency of the content. In the final stage, the feature grids are optimized to reconstruct the final image.
  }{
  }
  \label{fig:method_overview}
\end{figure*}

%% file: content/figures/rendering_overview.tex
\begin{figure}
\centering
\includegraphics[width=1.0\linewidth]{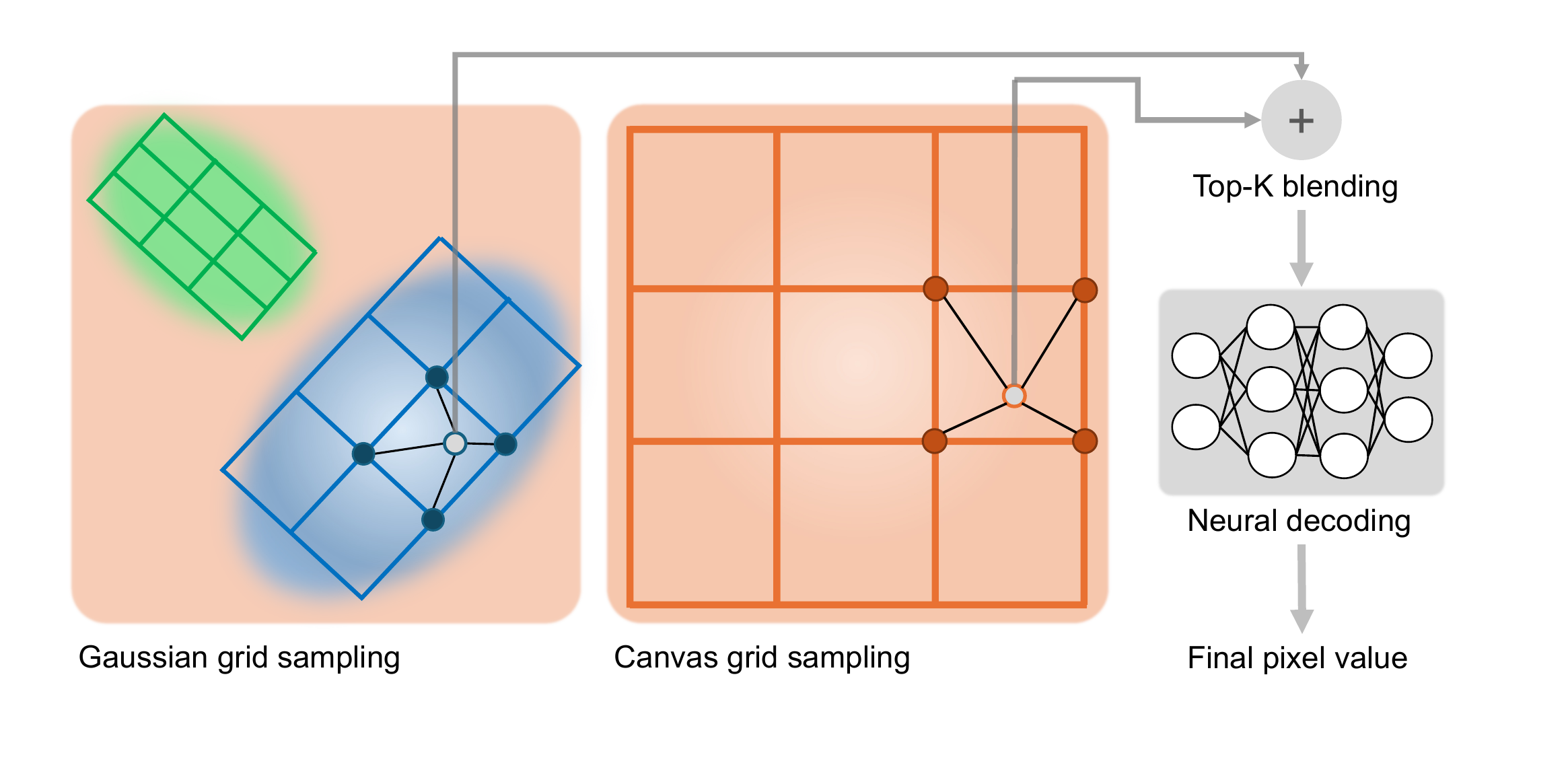}
\caption{Pixel Rendering. For a given pixel location, \projName bilinearly samples top-K overlapping Gaussian grids as well as the canvas grid. The interpolated features are decoded by a small \Decoder to render the final pixel value.}{}
\label{fig:rendering_overview}
\end{figure}

%% file: content/main/4-result.tex
\section{Evaluation}
\label{sec:result}

\subsection{Experimental setup}
\paragraph{Datasets}
We evaluate \projName's performance on the set of 100 images acquired from a validation subset of the DIV2K natural images dataset \cite{Agustsson2017DIV2K}. We provide an additional evaluation on the professional validation subset of the CLIC2020 benchmark \cite{CLIC2020}, 41 images in total.

\paragraph{Metrics and baselines}
Following the methodology of prior studies \cite{zhang2025image}, we evaluate reconstruction quality across various memory budgets, quantified in bits-per-pixel (bpp). We assess quality using standard error-based and perceptual metrics: PSNR, SSIM \cite{wang2004ssim}, MS-SSIM, and LPIPS \cite{zhang2018perceptual}. We compare \projName to three recent primitive-based methods, \cite{zhang2025image} (\textbf{IMG-GS}), \cite{zhang2024gaussianimage} (\textbf{GI}), and \cite{iinbor2026sad} (\textbf{SAD}). Additionally, we include a popular structured grid-based baseline, \cite{muller2022instant} (\textbf{I-NGP}).

\subsection{Reconstruction Performance}
As shown in \Cref{tab:baselines}, \projName achieves competitive results across all budget points on both DIV2K and CLIC2020. At the lowest budget (0.1\,bpp), our method leads all baselines on PSNR by a clear margin, reaching 25.58\,dB on DIV2K (+0.86\,dB over the next-best SAD) and 27.70\,dB on CLIC2020 (+0.75\,dB), with LPIPS remaining competitive with IMG-GS at the lowest budget. At 0.4\,bpp we match SAD on PSNR (29.36\,dB) while outperforming all methods on SSIM, MS-SSIM, and LPIPS, demonstrating that the quality gain is not limited to pixel fidelity. At higher budgets (0.7--1.0\,bpp) our method remains best or near-best across all metrics, achieving 32.55\,dB PSNR and an LPIPS of 0.1167 at 1.0\,bpp on DIV2K — a 0.12\,dB gain over SAD and an LPIPS reduction of 0.069 over the next-best GaussianImage. Results on CLIC2020 follow the same trend, with \projName ranking
first on almost all four metrics at every budget point.

The quantitative improvements are reflected qualitatively in~\Cref{fig:qualitative_comparison} at 0.1\,bpp. Perceptually, our method combines the adaptivity of unstructured methods such as IMG-GS without the ``oil painting'' artifacts that appear in SAD, IMG-GS, and GI in regions of missing detail. This is due to our method's use of local feature grids within each Gaussian footprint, which faithfully reproduces spatially coherent structure rather than relying solely on primitive color and opacity. I-NGP, by contrast, produces noise-like patterns at low budgets owing to hash-grid collisions. We provide additional qualitative examples in the supplemental material.

\input{content/figures/qualitative_comparison}

\input{content/tables/results-imggs}

\subsection{Ablation Study}
\label{sec:ablation}
We ablate our method through two complementary studies summarized in~\Cref{tab:ablation_arch}. The architecture ablation evaluates each component incrementally, starting from a structured uniform baseline constructed by skipping the 2D Gaussian optimization and placing a single uniform canvas grid over the entire image with 100\% of the budget. All hyperparameters like feature grid dimension, \Decoder parameters, and quantization are held fixed throughout; bit budget formerly split between Gaussians and canvas is consolidated into one uniform grid. Adding frequency-guided placement alone \emph{reduces} performance (-1.22 dB PSNR), since each Gaussian receives equal grid budget regardless of its local frequency content, leaving low-frequency regions underserved. Budget weighting, which allocates grid resolution proportionally to each Gaussian's frequency energy, recovers this loss and surpasses the previous condition by +1.39 dB. Directional grid stretching along the dominant local frequency orientation contributes a further modest gain (+0.04 dB). The full method additionally applies adaptive canvas budget modulation and Gaussian pruning on a per-image basis, yielding +0.47 dB PSNR over the uniform baseline.

Notably, we observe that some numerical metrics (e.g., LPIPS) are marginally worse with our method than with the uniform placement condition. In fact, frequency-guided placement is designed to reallocate representation capacity from flat, low-frequency regions toward high-frequency textured regions, \textit{thus preserving fine-grained details}. 
This trade-off is the source of the perceptual metric gap in~\Cref{tab:ablation_arch}. 
To further demonstrate the observation,
\Cref{fig:abl_example} shows a qualitative comparison. In the textured crop, our method improves over the uniform baseline on every metric including LPIPS, confirming that the reallocation succeeds where it is intended to. In the flat crop, canvas-only coverage introduces a slightly blurrier output, penalized by the perceptual metrics, which are sensitive to low-level texture statistics in smooth regions.
We further discuss the limitation and future work for perceptually-aware budget allocation in \Cref{sec:future-work}.

The loss ablation isolates the contribution of the Step 1 overlap penalty. Training with $\mathcal{L}_{\mathrm{L1}}$ alone produces placements with redundant, overlapping Gaussians that waste budget; adding $\mathcal{L}_{\mathrm{ov}}$ penalizes this behavior and improves PSNR by +0.24 dB while simultaneously improving all perceptual metrics, confirming that better spatial coverage in Step 1 translates directly to reconstruction quality in Step 2.

\input{content/tables/ablation}

\section{Application: GPU-Native Dataset Archiver}
\label{sec:archiver}
Modern computer vision pipelines are fundamentally GPU-bound: training and inferring neural network models all require images loaded to GPU memory with local access for each GPU core. 
However, before images reach GPUs, our common practice is to unzip a large image dataset to the hard drive, decode the encoded image format (e.g., JPEG) into a decompressed pixel grid, and have them traverse the PCIe bus. 
Our neural image representation sidesteps this time-consuming bottleneck. Rather than the CPU-dependent data decoding to pixel values, our model-represented images can be loaded directly on the GPU. 
To demonstrate the application, we test the decoding efficiency of our method against JPEG \cite{wallace1991jpeg} + ZIP combination on the DIV2K validation set across four budget targets. This setup closely resembles image dataset storage pipelines. All decode timings cover the full pipeline from a compressed file in the OS page cache: our path performs loading of the quantized file, GPU dequantization, and CUDA kernel execution; the JPEG path amortizes a one-time ZIP extraction (87--130\,ms total, 0.87--1.30\,ms per image) and then performs per-image JPEG decode and transfer to GPU. Both paths are benchmarked as the mean over three synchronized iterations after two warmup passes, with \texttt{torch.cuda.synchronize()} enforcing completion. We conduct all comparisons on NVIDIA RTX 5080 GPU. We summarize the results in~\Cref{tab:full_comparison}. At 0.1 bpp, our method achieves $+2.33$\,dB higher PSNR than size-matched JPEG ($\approx47$\,KB per image), decodes in $5.44$\,ms versus $11.65$\,ms ($2.14\times$ faster), and transfers $47$\,KB over PCIe versus $8{,}306$\,KB of decoded pixels, a $176\times$ reduction in bus traffic. At 0.4 bpp the quality gap narrows to $+0.31$\,dB in our favour; above 0.4 bpp the advantage reverses ($-0.53$ to $-1.00$\,dB at 0.7--1.0\,bpp) as JPEG's entropy coder becomes more competitive at higher budgets. The decode speedup persists across all operating points ($1.24$--$2.14\times$), and the PCIe reduction remains substantial ($23\times$--$176\times$) at all budgets. This is not a replacement for JPEG in general-purpose imaging; it targets pipelines where images are consumed exclusively on GPU, PCIe is a shared resource across concurrent tasks, and the 0.1 bpp regime is appropriate, such as large-scale retrieval, streaming previews, and thumbnail caching.

\input{content/tables/performance}

%% file: content/figures/qualitative_comparison.tex
\begin{figure*}
\centering
  \subfloat[I-NGP]{
    \includegraphics[width=0.1\linewidth]{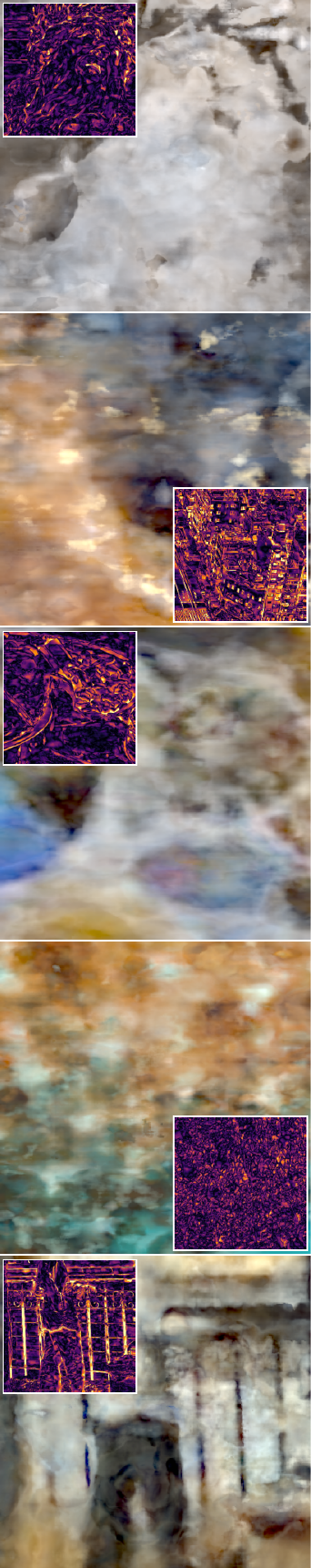}\hspace{-0.4em}
  } 
    \subfloat[GI]{
    \includegraphics[width=0.1\linewidth]{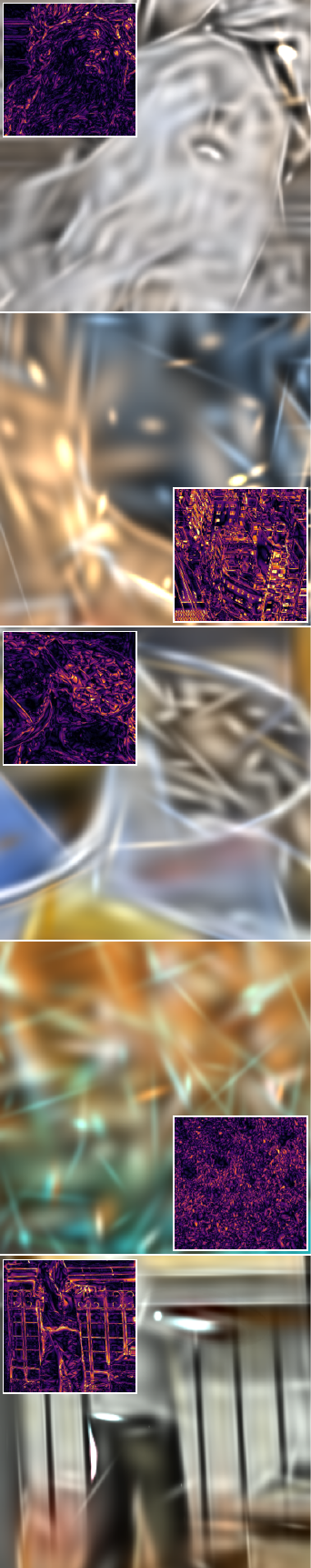}\hspace{-0.4em}
  }
  \subfloat[SAD]{
    \includegraphics[width=0.1\linewidth]{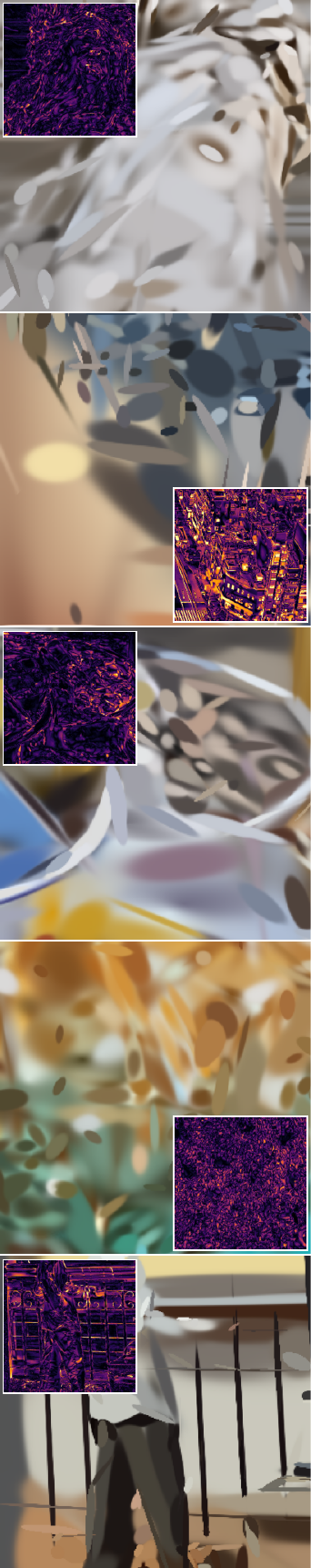}\hspace{-0.4em}
  }
  \subfloat[IMG-GS]{
    \includegraphics[width=0.1\linewidth]{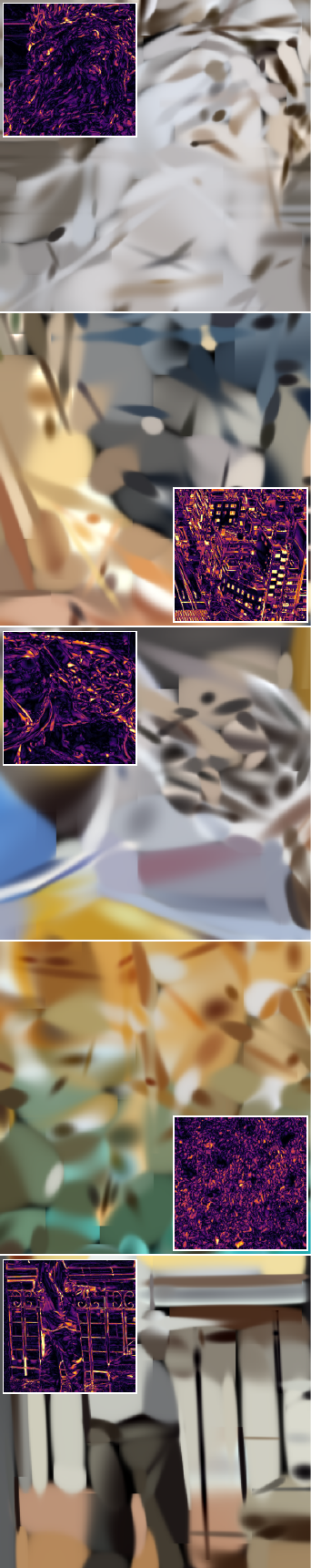}\hspace{-0.4em}
  }
  \subfloat[Ours]{
    \includegraphics[width=0.1\linewidth]{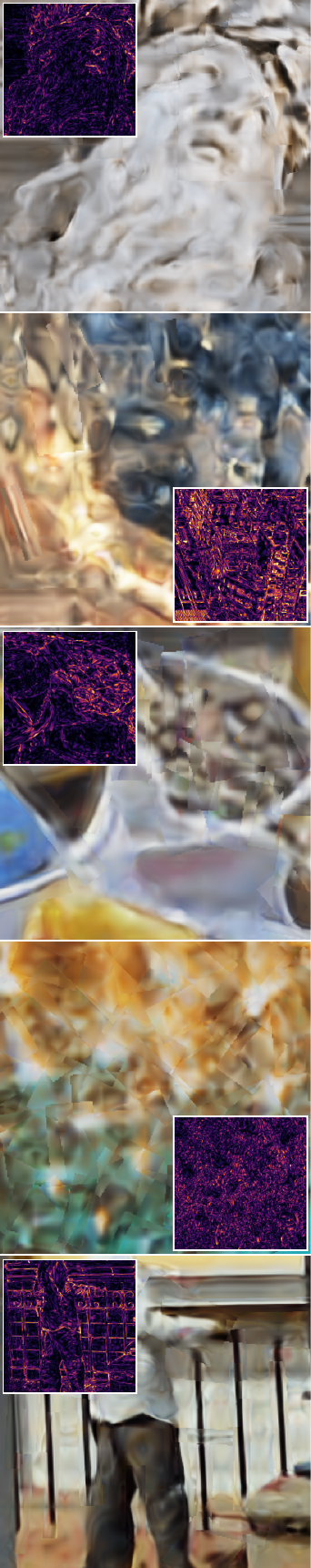}\hspace{-0.4em}
  }
  \subfloat[Ref. Crop]{
    \includegraphics[width=0.1\linewidth]{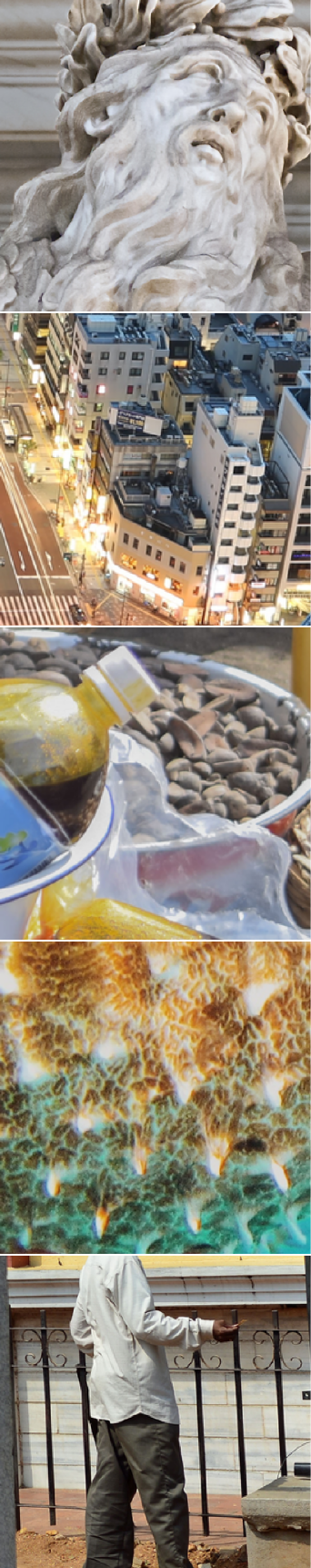}\hspace{-0.4em}
  }
  \subfloat[Reference]{
    \includegraphics[width=0.11\linewidth]{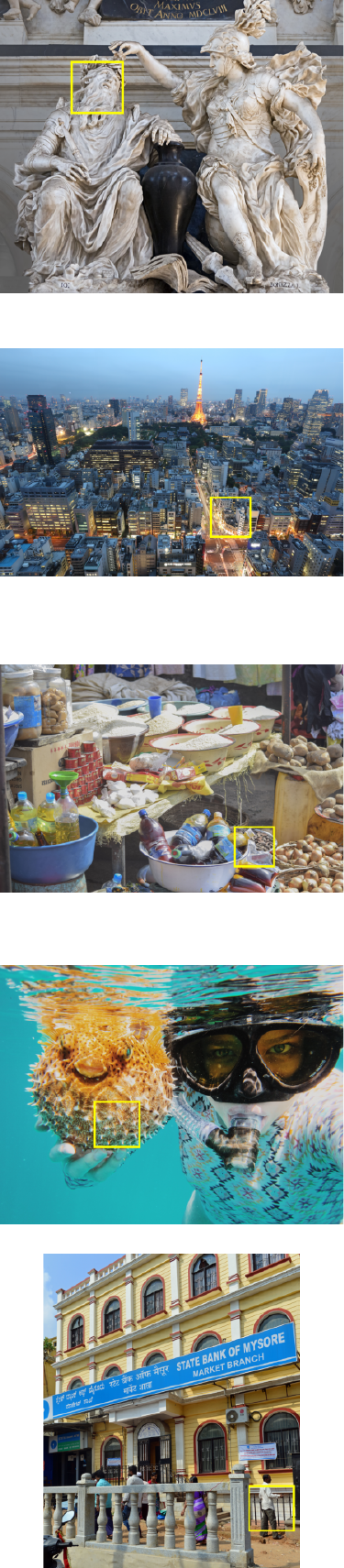}\hspace{-0.4em}
  }
  \caption{Qualitative comparison with other recent image representation baselines. Examples are from the DIV2K dataset. Columns (a)-(e) show model reconstructions with error map insets, (f) and (g) show reference crops and the full reference images respectively. The budgets of I-NGP, GI, SAD, IMG-GS, and \projName are 0.1 bits per pixel respectively.}{}
  \label{fig:qualitative_comparison}
\end{figure*}

%% file: content/tables/results-imggs.tex
\begin{table*}
    \centering
    \caption{Comparison of \projName with recent primitive-based and neural baselines on DIV2K (100 images) and CLIC2020 validation (41 images) natural image datasets across four budgets (0.1 bits per pixel to 1.0 bits per pixel). The best result is in dark blue bold and the second best is in light blue.}{}
    \label{tab:baselines}
    \footnotesize
    \begin{tabular}{c|c|cccc|cccc}
        \toprule
        & & \multicolumn{4}{c|}{DIV2K} & \multicolumn{4}{c}{CLIC2020} \\
        Method & Metric & 0.1\,bpp & 0.4\,bpp  & 0.7\,bpp  & 1.0\,bpp  & 0.1\,bpp  & 0.4\,bpp  & 0.7\,bpp  & 1.0\,bpp  \\
        \midrule
        \multirow{4}{*}{SAD \cite{iinbor2026sad}}
            & PSNR $\uparrow$ & 24.72 \second & \first{29.36} & \first{31.24} & 32.43 \second & 26.95 \second & 31.29 \second & 33.04 \second & 34.10 \second \\
            & SSIM $\uparrow$ & 0.6573 & 0.7816 \second & 0.8309 \second & 0.8597 \second & 0.7210 & 0.8096 & 0.8465 \second & 0.8698 \second \\
            & MS-SSIM $\uparrow$ & 0.7954 & 0.9196 & 0.9495 \second & 0.9632 \second & 0.8353 & 0.9271 & 0.9518 & 0.9643 \second \\
            & LPIPS $\downarrow$ & 0.4512 & 0.2980 & 0.2399 & 0.2040 & 0.4238 & 0.2996 & 0.2524 & 0.2197 \\
            \midrule
        \multirow{4}{*}{GI \cite{zhang2024gaussianimage}}
            & PSNR $\uparrow$ & 24.21 & 27.27 & 28.71 & 29.69 & 26.36 & 29.15 & 30.40 & 31.33 \\
            & SSIM $\uparrow$ & 0.6512 & 0.7640 & 0.8121 & 0.8402 & 0.7147 & 0.7980 & 0.8308 & 0.8528 \\
            & MS-SSIM $\uparrow$ & 0.8398 \second & 0.9271 \second & 0.9488 & 0.9588 & 0.8707 \second & 0.9364 \second & 0.9527 & 0.9613 \\
            & LPIPS $\downarrow$ & 0.4408 & 0.2812 & 0.2247 & 0.1856 \second & 0.3718 & 0.2546 & 0.2103 & 0.1778 \second \\
            \midrule
        \multirow{4}{*}{I-NGP \cite{muller2022instant}}
            & PSNR $\uparrow$ & 22.16 & 25.30 & 26.27 & 26.98 & 24.85 & 27.55 & 28.38 & 28.96 \\
            & SSIM $\uparrow$ & 0.5341 & 0.6656 & 0.7152 & 0.7501 & 0.6396 & 0.7281 & 0.7637 & 0.7894 \\
            & MS-SSIM $\uparrow$ & 0.7163 & 0.8602 & 0.8963 & 0.9184 & 0.7951 & 0.8888 & 0.9154 & 0.9308 \\
            & LPIPS $\downarrow$ & 0.5435 & 0.3749 & 0.3109 & 0.2599 & 0.4755 & 0.3344 & 0.2807 & 0.2447 \\
            \midrule
        \multirow{4}{*}{IMG-GS \cite{zhang2025image}}
            & PSNR $\uparrow$ & 24.46 & 27.97 \second & 29.49 & 30.51 & 26.66 & 30.15 & 31.66 & 32.62 \\
            & SSIM $\uparrow$ & 0.6711 \second & 0.7754 & 0.8180 & 0.8438 & 0.7329 \second & 0.8112 \second & 0.8423 & 0.8607 \\
            & MS-SSIM $\uparrow$ & 0.8243 & 0.9218 & 0.9467 & 0.9588 & 0.8586 & 0.9346 & 0.9536 \second & 0.9628 \\
            & LPIPS $\downarrow$ & \first{0.3891} & 0.2682 \second & 0.2217 \second & 0.1927 & \first{0.3370} & 0.2395 \second & 0.2048 \second & 0.1834 \\
            \midrule
        \multirow{4}{*}{Ours}
            & PSNR $\uparrow$ & \first{25.58} & \first{29.36} & 31.22 \second & \first{32.55} & \first{27.70} & \first{31.32} & \first{33.06} & \first{34.26} \\
            & SSIM $\uparrow$ & \first{0.6997} & \first{0.8253} & \first{0.8730} & \first{0.8999} & \first{0.7523} & \first{0.8448} & \first{0.8821} & \first{0.9039} \\
            & MS-SSIM $\uparrow$ & \first{0.8869} & \first{0.9554} & \first{0.9713} & \first{0.9787} & \first{0.9074} & \first{0.9604} & \first{0.9735} & \first{0.9798} \\
            & LPIPS $\downarrow$ & 0.3902 \second & \first{0.2178} & \first{0.1545} & \first{0.1167} & 0.3601 \second & \first{0.2210} & \first{0.1665} & \first{0.1324} \\
        \bottomrule
    \end{tabular}
\end{table*}

%% file: content/tables/ablation.tex
\begin{figure*}[t]
  \begin{minipage}[c]{0.45\textwidth}
    \captionof{table}{Step\,1 loss and Step\, 2 architecture ablations at 1.0\,bpp on 100 DIV2K
      validation images. For Step\,1 loss ablation, the architecture is fixed to the
      full method. Each row adds one component to the previous. The best result is in dark blue bold and the second best is in light blue. We further clarify the perceptual metric result in~\Cref{sec:ablation} and in~\Cref{fig:abl_example}.}
    \label{tab:ablation_arch}
    \centering
    \setlength{\tabcolsep}{3pt}
    \footnotesize
    \begin{tabular}{lcccc}
      \toprule
      Step\,1 Conditions & PSNR\,$\uparrow$ & SSIM\,$\uparrow$ & MS-SSIM\,$\uparrow$ & LPIPS\,$\downarrow$ \\
      \midrule
      $\mathcal{L}_{\mathrm{L1}}$                                          & 32.31 \second  & 0.8852 \second & 0.9752 \second & 0.1434 \second \\
      $\mathcal{L}_{\mathrm{L1}} + \mathcal{L}_{\mathrm{ov}}$ (full) & \first{32.55}  & \first{0.8999} & \first{0.9787} & \first{0.1167} \\
      \midrule
      Step\,2 Conditions & \\
      \midrule
      Uniform placement        & 32.08         & \first{0.9069} & \first{0.9809} & \first{0.0993} \\
      + Freq.-guided placement & 30.86         & 0.8775         & 0.9738         & 0.1399         \\
      + Budget weighting       & 32.25         & 0.8844         & 0.9750         & 0.1437         \\
      + Directional grids      & 32.29 \second & 0.8866         & 0.9756         & 0.1413         \\
      Ours (full)              & \first{32.55} & 0.8999 \second & 0.9787 \second & 0.1167 \second \\
      \bottomrule
    \end{tabular}
  \end{minipage}
  \hfill
  \begin{minipage}[c]{0.52\textwidth}
    \centering
    \includegraphics[width=\linewidth]{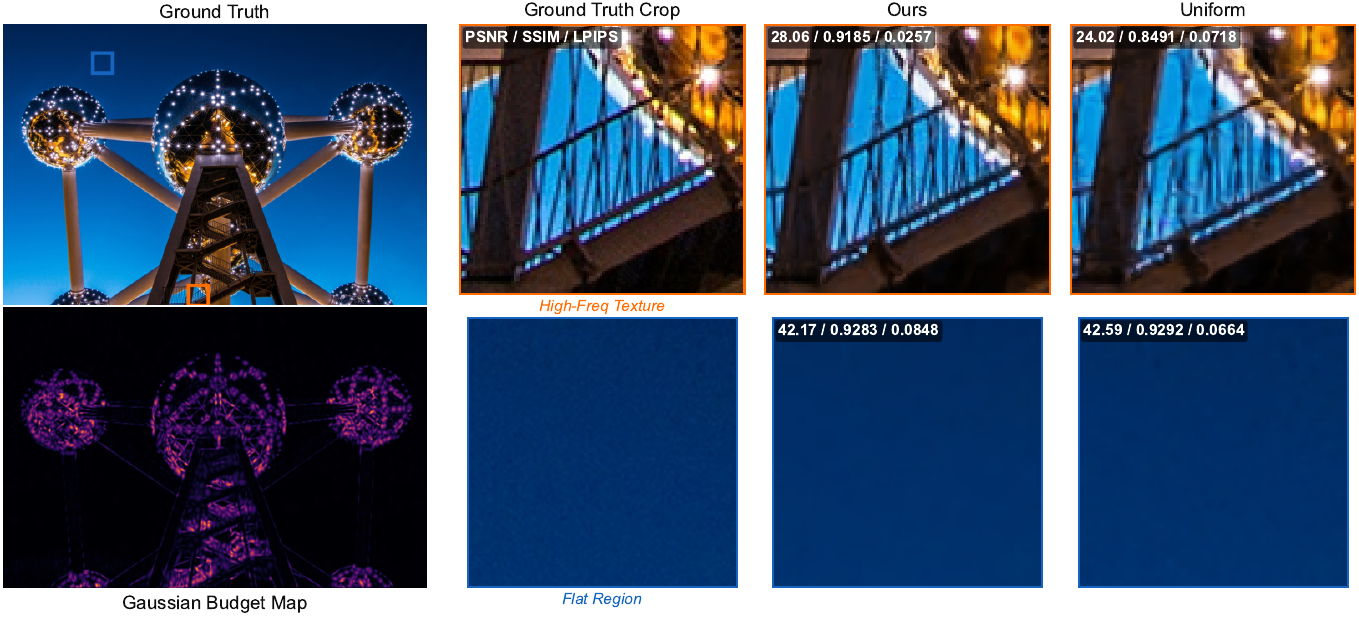}
    \captionof{figure}{\protect Qualitative comparison between our full method and the uniform placement baseline on DIV2K image 0851 at 1.0 bpp. \textbf{Top row (textured region):} Frequency-guided Gaussian grids preserve complex details, yielding improvements across all three metrics (e.g., sharper details along the fence). \textbf{Bottom row (flat region):} Capacity reallocation leaves smooth areas with low-frequency, canvas-only coverage, resulting in marginally lower metrics yet visually negligible to observers. This demonstrates how our method prioritizes critical details at a minor cost to flat regions that is imperceptible.}
    \label{fig:abl_example}
  \end{minipage}
\end{figure*}

%% file: content/tables/performance.tex
\begin{table*}[t]
\centering
\setlength{\tabcolsep}{3pt}
\footnotesize
\caption{%
  End-to-end decode comparison against JPEG + ZIP on 100 DIV2K images
  at four budgets.
  At 0.1\,bpp our representation yields $+2.33$\,dB over a size-matched JPEG
  while decoding $2.14\times$ faster;
  The decode speedup ($1.24$--$2.14\times$, Speedup column) persists across
  all budgets. The rightmost column ($\text{PCIe}\times$) shows the most
  distinctive property: transferring quantized model weights (47--366\,KB)
  instead of decoded pixel maps ($8{,}306$\,KB per image) reduces
  CPU-to-GPU traffic by $23$--$176\times$.
  \textbf{Pipeline columns.}
  Ours: weights load (\emph{Load}), GPU transfer
  (\emph{PCIe}), GPU dequantize (\emph{Dq}), GPU kernel forward pass
  (\emph{Krnl}).
  JPEG: one-time ZIP extraction on CPU amortized over 100 images
  (\emph{Unzip}), JPEG CPU decode (\emph{Dec}), and pixel transfer to GPU
  (\emph{PCIe}).
}
\label{tab:full_comparison}
\resizebox{\textwidth}{!}{%
\begin{tabular}{r rr rrr rrrr rrrrr r r}
\toprule
 & \multicolumn{2}{c}{Dataset size (MB)}
 & \multicolumn{3}{c}{PSNR (dB)}
 & \multicolumn{4}{c}{JPEG decode (ms\,/\,image)}
 & \multicolumn{5}{c}{Our decode (ms\,/\,image)}
 & \multicolumn{1}{c}{}
 & \multicolumn{1}{c}{}
 \\
\cmidrule(lr){2-3}\cmidrule(lr){4-6}\cmidrule(lr){7-10}\cmidrule(lr){11-15}
bpp
  & JPEG & Ours
  & JPEG & Ours & $\Delta$
  & Unzip & Dec & PCIe & Total
  & Load & PCIe & Dq & Krnl & Total
  & Speedup
  & $\text{PCIe}\times$
\\
\midrule
0.1
  & 4.85  & 4.62
  & $23.25$ & $\mathbf{25.58}$ & $\mathbf{+2.33}$
  & 0.87 & 10.05 & 0.73 & 11.65
  & 0.17 & 0.14 & 0.64 & 4.48 & $\mathbf{5.44}$
  & $\mathbf{2.14\times}$
  & $176\times$ \\
0.4
  & 14.71 & 14.85
  & $29.05$ & $\mathbf{29.36}$ & $\mathbf{+0.31}$
  & 1.20 & 9.46 & 0.67 & 11.33
  & 0.16 & 0.15 & 0.63 & 6.78 & $\mathbf{7.73}$
  & $\mathbf{1.47\times}$
  & $55\times$ \\
0.7
  & 24.97 & 25.28
  & $\mathbf{31.75}$ & $31.22$ & $-0.53$
  & 1.26 & 9.88 & 0.64 & 11.78
  & 0.16 & 0.15 & 0.57 & 8.25 & $\mathbf{9.13}$
  & $\mathbf{1.29\times}$
  & $32\times$ \\
1.0
  & 35.48 & 35.71
  & $\mathbf{33.55}$ & $32.55$ & $-1.00$
  & 1.30 & 10.61 & 0.64 & 12.55
  & 0.16 & 0.16 & 0.59 & 9.19 & $\mathbf{10.11}$
  & $\mathbf{1.24\times}$
  & $23\times$ \\
\midrule
All
  & 80.0 & 80.5
  & \multicolumn{2}{c}{---} & $\mathbf{+0.27}$
  & 1.16 & 10.00 & 0.67 & 11.83
  & 0.16 & 0.15 & 0.61 & 7.18 & $\mathbf{8.10}$
  & $\mathbf{1.46\times}$
  & $40\times$ \\
\bottomrule
\end{tabular}}
\end{table*}

%% file: content/main/5-future-work.tex
\section{Limitations and Future Work}
\label{sec:future-work}
The gains of \projName{} depend on how spatial heterogeneity of an image
frequency content. When detail is concentrated, frequency-guided
allocation directs capacity to complex regions and yields clear gains; as
frequency energy becomes uniform, the allocation approaches the uniform
baseline (see supplement). Perceptual metrics do not always follow the PSNR
gains, since capacity is drawn away from smooth regions.
In the future, we plan to investigate human visual contrast sensitivity function based on eye-display distances  \cite{rovamo1994contrast} for our resource allocation weighing. Integrating the visual sensitivity to extend the currently high-frequency preservation would yield increased perceptual quality for human observers. 
Additionally, \projName encodes each image by per-image optimization, which takes on average 24.3\,min per DIV2K image on a single NVIDIA A100, almost all of it in Step~2. This is far slower than traditional codecs such as JPEG, and we do not claim competitive encoding times against the primitive-based baselines. Our target setting is write-once, read-many storage (\Cref{sec:archiver}). Shorter Step~2 schedules with early stopping, or learned initialization of the grid features, could reduce this cost and are left for future work.

\section{Conclusion}
In this work, we proposed \projName, a hybrid content-adaptive image representation utilizing frequency-guided feature grids via 2D Gaussian primitives. \projName achieves a favorable quality at low budgets, while preserving a significant perceptual win and fast rendering speeds, making it suitable for GPU-native data storage. We hope this work inspires future advances in content-adaptive image representations.